\documentclass[11pt]{article}

\usepackage{coling2018}
\usepackage{times}
\usepackage{url}
\usepackage{latexsym}
\usepackage{multirow}
\usepackage{amsmath}
\usepackage{siunitx}
\usepackage{color}
\usepackage{graphicx}
\usepackage{subcaption}

\def\FromEn{\texttt{<FromEn>}}
\def\FromFr{\texttt{<FromFr>}}
\def\FromEs{\texttt{<FromEs>}}
\def\FromDe{\texttt{<FromDe>}}

\def\ToEn{\texttt{<ToEn>}}
\def\ToFr{\texttt{<ToFr>}}
\def\ToEs{\texttt{<ToEs>}}
\def\ToDe{\texttt{<ToDe>}}

\def\FrEn{\texttt{<FrEn>}}
\def\EnFr{\texttt{<EnFr>}}
\def\EsEn{\texttt{<EsEn>}}
\def\EnEs{\texttt{<EnEs>}}
\def\DeEn{\texttt{<DeEn>}}
\def\EnDe{\texttt{<EnDe>}}

\def\xxen{xx-to-En}
\def\enxx{En-to-xx}

\title{Multilingual Neural Machine Translation with Task-Specific Attention}

\author{Graeme Blackwood ~~~~~~ Miguel Ballesteros ~~~~~~ Todd Ward \\
  IBM Research AI \\
  1101 Kitchawan Rd, Yorktown Heights, NY 10598, USA\\
  {\tt \{blackwood,toddward\}@us.ibm.com, miguel.ballesteros@ibm.com} \\}

\date{}

\begin{document}
\maketitle
\begin{abstract}
Multilingual machine translation addresses the task of translating between multiple source and target languages. We propose task-specific attention models, a simple but effective technique for improving the quality of sequence-to-sequence neural multilingual translation. Our approach seeks to retain as much of the parameter sharing generalization of NMT models as possible, while still allowing for language-specific specialization of the attention model to a particular language-pair or task. Our experiments on four languages of the Europarl corpus show that using a target-specific model of attention provides consistent gains in translation quality for all possible translation directions, compared to a model in which all parameters are shared. We observe improved translation quality even in the (extreme) low-resource zero-shot translation directions for which the model never saw explicitly paired parallel data.
\end{abstract}

\blfootnote{
\hspace{-0.65cm} 
This work is licensed under a Creative Commons Attribution 4.0 International License.
License details: \\
\url{http://creativecommons.org/licenses/by/4.0/}.
}

\section{Introduction and Motivation}
\label{sec:introduction}
Multilingual machine translation is the task of building a system capable of translating between more than just a single source and target language. The approach is motivated by the idea that learning to translate between one pair of languages can help to improve the translation quality of related language pairs. Multilingual MT can be divided into three main types according to the support for source and target languages: (i) single source, multiple target (ii) multiple source, single target, and (iii) multiple source, multiple target. Our work focuses on improving the quality of fully $n$-way multilingual NMT models that can translate between multiple source and target languages.

Multinational organizations and companies increasingly publish content in a variety of languages. In addition to exploiting multiple sources of parallel data and leveraging similarities across different languages, multilingual translation can considerably simplify deployment. Directly supporting all possible translation directions for a set of $n$ languages would require $n(n-1)$ sets of parallel data and trained models. This is an expensive proposal, from both a data and deployment perspective. The number of required systems can be reduced to $n$ if translation by bridging is used. In contrast, a fully $n$-way multilingual system can be trained to translate between all known language pairs using a single model. It is even possible to translate between language pairs that were never explicitly paired in the parallel training data, so called zero-shot translation \cite{TACL1081}.

Neural machine translation (NMT) \cite{DBLP:journals/corr/BahdanauCB14,DBLP:conf/nips/SutskeverVL14} has recently become the standard model of translation due to the high levels of fluency and accuracy that it can achieve \cite{DBLP:journals/corr/KoehnK17}. NMT is an excellent choice for multilingual translation since the neural architecture is language-agnostic and capable of capturing translation properties, such as long-distance re-ordering, between even highly dissimilar languages. One of the main advantages of NMT is that the model is able to learn useful linguistic representations for many languages, and to share parameters and leverage similarities in the abstractions learned by the embeddings and hidden layers of the model. Exploiting multilingual data and representations is of particular interest in improving the quality of translation for low-resource (or even zero-resource) language pairs.

In this paper, we describe a simple but effective extension to the attentional model of neural machine translation that improves the quality of multilingual NMT. We seek to retain as much of the parameter sharing of NMT models as possible, while allowing for language-specific specialization of the attention model to a particular task. Our experiments on the Europarl corpus show that multilingual translation quality is improved for all tested language pairs, with the largest improvements in the (extreme low-resource) zero-shot directions.

The architecture of our sequence-to-sequence neural machine translation system is described in Section \ref{sec:architecture}. An overview of related work in multilingual machine translation and multi-task learning is provided in Section \ref{sec:related-work}. In Section \ref{sec:task-specific-attention}, we introduce three task-specific attention model variants that can be used to improve the quality of multilingual NMT. Our experimental framework, implementation and results are described in Section \ref{sec:experiments}, followed by an analysis and example translations with attention plots in Section \ref{sec:analysis-examples}. We conclude with a discussion of possible future work in Section \ref{sec:conclusions-future-work}.

\section{Sequence to Sequence Translation with Attention}
\label{sec:architecture}
Our NMT system is based on the sequence-to-sequence model of translation described by \newcite{DBLP:journals/corr/BahdanauCB14}, consisting of a recurrent neural network encoder and decoder with an attention mechanism.


The encoder uses bi-directional gated recurrent units (GRU) \cite{DBLP:journals/corr/ChoMBB14} to encode a source sequence ${\bf{x}}=(x_1,\ldots,x_l)$, where $x_i$ is the embedding vector for the $i$-th word of the source sentence and $l$ its length. The encoded form of the complete sentence is defined by the sequence of hidden states ${\bf{h}} = (h_1,\ldots,h_l)$, where each $h_i$ is computed as
\begin{equation}
h_i = 
\begin{bmatrix}
\overleftarrow{h}_i \\
\overrightarrow{h}_i
\end{bmatrix}
=
\begin{bmatrix}
\overleftarrow{f}(x_i, \overleftarrow{h}_{i+1}) \\
\overrightarrow{f}(x_i, \overrightarrow{h}_{i-1})
\end{bmatrix},
\end{equation}
and $\overleftarrow{f}$ and $\overrightarrow{f}$ denote the operations of the right-to-left and left-to-right GRU cells, respectively.

Given the encoded representation of the source sentence $\bf{h}$, the decoder produces the target translation $\mathbf{y}$ by computing the sequence $\mathbf{y} = (y_1,\ldots,y_m)$, where $m$ is the target sentence length. At each time-step $t$, the probability of target word $y_t$ is computed as
\begin{equation}
p(y_t|\mathbf{h},y_{t-1},\ldots,y_1) = g(s_t, y_{t-1}, H_t),
\end{equation}
where $g$ is a feed-forward network over the current decoder hidden state $s_t$, the word embedding of the previously predicted target word $y_{t-1}$, and a context vector $H_t$, followed by a softmax to predict the probability distribution over the output vocabulary. The decoder states are updated according to
\begin{equation}
s_t = q(s_{t-1},y_{t-1},H_t),
\end{equation}
where $q$ implements the conditional GRU with attention of \newcite{sennrich-EtAl:2017:EACLDemo} and $H_t$ is an attention-weighted representation of the encoded source sequence $\mathbf{h}$. The attention-weighted source representation at time-step $t$ is defined as the weighted sum
\begin{equation}
H_t =
\begin{bmatrix}
\sum^l_{i=1} (\alpha_{t,i} \cdot \overleftarrow{h}_i) \\
\sum^l_{i=1} (\alpha_{t,i} \cdot \overrightarrow{h}_i)
\end{bmatrix},
\end{equation}
where the normalized weights $\alpha_{t,i}$ indicate the relative importance of the vector representation of each source position $h_i$ when producing the next target word at time-step $t$. The weights are computed using a two-layer feed-forward network $r$:
\begin{equation}
\alpha_{t,i} = \frac{\exp\{r(s_{t-1},h_i,y_{t-1})\}}{\sum_k \exp\{r(s_{t-1},h_k,y_{t-1})\}}.
\label{eqn:attention-model}
\end{equation}

\section{Related Work}
\label{sec:related-work}
Early phrase-based approaches to multilingual MT focused on multi-source translation. \newcite{franz-ney-2001:multi-source} used a simple product or max rule to select at the sentence-level the single best hypothesis with the highest translation score from multiple decoders. There is no sharing of parameters. Consensus network decoding \cite{matusov:2006} can also be used to combine the word-level output of translations from multiple source languages. Such system combination techniques allow sharing of the words in the candidate translations, but still require training individual models for each language pair of interest.

Training multilingual MT systems capable of translating between multiple languages can be considered an instance of multi-task learning \cite{DBLP:journals/ml/Caruana97}, which is the idea of solving synergistic tasks while maximizing the number of shared parameters. Sharing parameters may be useful when attempting to solve different tasks, since we can minimize representation bias by learning a more regularized representation \cite{DBLP:journals/jair/Baxter00}. The flexibility nature of neural architectures allows for selection of the components of the model that are to be shared and those which are not.

Sequence to sequence models with attention are no exception. Each set of parameters provides different levels of generalization \cite{DBLP:conf/emnlp/ReimersG17}, which is evidenced in the synergistic task of training multilingual translation models. For example, \newcite{dong-EtAl:2015:ACL-IJCNLP2} jointly train decoders while the rest of the parameters are task-specific; \newcite{zoph-knight:2016:N16-1} jointly train the encoders while the rest of the parameters are task-specific, and \newcite{TACL1081} train both encoders and decoders jointly with language-specific tokens to guide learning as in \cite{DBLP:journals/tacl/AmmarMBDS16}. These latter approaches are the ones that we build on. We augment our decoder with a task-specific attention mechanism intended to better capture word order and language-specific nuances while continuing to share the rest of the model parameters (including token embeddings).

\section{Task-Specific Attention Models}
\label{sec:task-specific-attention}
Fully $n$-way multilingual NMT systems need to support multiple source and target languages in the encoder and decoder GRUs. Our work focuses on improving the use of attention in the decoder. At each time-step $t$, the decoder computes the attention model weights $\alpha_{t,i}$ of Equation~(\ref{eqn:attention-model}) in order to quantify the relative importance of the encoder states corresponding to each source position. To the extent that attention can be considered an analogue of word alignment, we would expect to see different patterns of attention for language-pairs with different word order and word-level alignments. Multilingual models restricted to a single shared attention will struggle when decoding multiple languages with different word orders, since the parts of the source sentence that should be attended to depend on the source and target languages of the translation task.

Our task-specific attention models can be used to address this issue. We train and empirically evaluate three task-specific attention model variants: \textit{target-specific} attention, \textit{source-specific} attention, and \textit{paired} attention which is associated with a particular language pair (i.e.\ translation direction). Each of our models introduces conditioning on the weights and biases used to compute the attention coefficients. They differ only in the choice of the key used for conditioning attention:

\begin{itemize}
\item \textit{Target-specific}: separate attention weights and bias for each target language
\item \textit{Source-specific}: separate attention weights and bias for each source language
\item \textit{Paired}: separate attention weights and bias for each source + target pair
\end{itemize}

Our multilingual NMT system follows \newcite{TACL1081} in using special tokens to indicate the desired target language. These tokens can be considered to define the `task' from a multi-task learning perspective as shown in \cite{DBLP:journals/tacl/AmmarMBDS16,N18-1005}. Since the encoder states are composed from bi-directional GRUs, we augment the source side of our parallel training data with both prefix and suffix task tokens. This ensures that the task token is not attenuated in the left-to-right encoder GRU. Such tokens are sufficient to ensure that the multilingual decoder produces words in the correct target language\footnote{We tried explicit softmax decoding constraints on the target vocabulary, but found them to be unnecessary.}. A French source sentence that is to be translated into English would be augmented as follows:

\begin{center}
\ToEn\ Guide des industries canadiennes : \ToEn\
\end{center}

During training and decoding, we dynamically construct the computation graph using the parameters for each task. Table \ref{tab:valid-prefix-tokens} defines the valid task tokens for target-specific, source-specific and paired attention models, given the four languages supported by our multilingual model. See Section \ref{sec:experiments} for full details of our experimental framework and supported language pairs.

\begin{table}[t]
\begin{center}
\begin{tabular}{r|l}
target-specific & \ToEn, \ToFr, \ToEs, \ToDe \\
source-specific & \FromEn, \FromFr, \FromEs, \FromDe \\
paired          & \FrEn, \EnFr, \EsEn, \EnEs, \DeEn, \EnDe \\
\end{tabular}
\end{center}
\caption{Valid source-side task tokens for task-specific attention model training and decoding.}
\label{tab:valid-prefix-tokens}
\end{table}

The augmented source sides of the parallel training data for our three model variants take the abstract forms shown in Table \ref{tab:augmented-sources}. The target-specific and paired attention models simply add the desired target language or language pair prefix and suffix tokens. For the source-specific attention model, we want to continue to allow for run-time selection of the desired target language so we introduce a second prefix token to indicate the selection of the specific attention parameters. The first token for the source-specific models is used only to select the parameters and stripped from the source sentence before using it in training or decoding. This allows us to dynamically select the source-specific attention parameters associated with translation from French, while (i) still allowing run-time selection of the target language, (ii) supporting zero-shot directions, and (iii) ensuring that the multilingual model sees exactly the same sequence of tokens as the target-specific model. 

\begin{table}[t]
\begin{center}
\begin{tabular}{r|l}
target-specific & \ToFr\ $w_1$ $w_2$ \ldots $w_l$ \ToFr\ \\
source-specific & \FromFr\ \ToEn\ $w_1$ $w_2$ \ldots $w_l$ \ToEn\ \\
paired          & \FrEn\ $w_1$ $w_2$ \ldots $w_l$ \FrEn\ \\
\end{tabular}
\end{center}
\caption{Source side training data augmented with task-specific attention model tokens.}
\label{tab:augmented-sources}
\end{table}

Note that only the target-specific and source-specific model variants enable zero-shot translation. The paired form of attention model trains separate attention weights and biases for each of the translation directions observed in the training data. It would be possible to use a separate prefix token for the attention key (in a similar manner to that of source-specific attention), but there is no explicit set of attention parameters that should be used for the zero-shot directions. The prefix token could specify that the attention parameters associated with either the source or target language of the zero-shot direction should be used.

All three of our task-specific attention model variants still share most of their encoder and decoder parameters. The task-specific attention models require only a very small increase in the total number of parameters. For hidden state dimensionality $d$, we add one additional set of attention weights ($d \times d$ parameters) and bias ($d$ parameters) for each supported task. For the neural network topology used in our experiments (see Section \ref{sec:experiments}, below), a target-specific attention model with support for four distinct target languages (i.e.\ tasks) requires only a 1.2\% increase in the total number of model parameters, compared to the shared-attention version of the model.


\section{Experiments}
\label{sec:experiments}
We evaluate the quality of our multilingual translation models using training data from the Europarl Corpus\footnote{\url{http://www.statmt.org/europarl/}}, Release V7. Our experiments focus on three primary language pairs: French-English, Spanish-English and German-English. We include German with SOV-type word order to contrast the SVO-type word order of the other languages. The source and target sides of the parallel training data are processed with the standard tokenizer included in the Moses SMT Toolkit \cite{koehn:moses:07}. For training multilingual systems, we merge the parallel data for all available directions. The parallel data for each language pair is thus included twice, once in each direction.

In order to support experiments on many-to-many multilingual translation using a single model, we apply a jointly-learned set of 80k Byte-Pair Encoding (BPE) rules \cite{DBLP:journals/corr/SennrichHB15} obtained from the merged source and target sides of the training data for all three language pairs. This ensures that all experiments, including the single-language baselines, share exactly the same tokenization and sub-word vocabularies. Tokenized corpus statistics for the parallel training data are summarized in Table~\ref{table:tok-corpus-stats}, while Table \ref{table:voc-sizes-bpe} compares the vocabulary sizes obtained from BPE processed data for the single-data baseline and fully $n$-way multilingual models.

\begin{table}[t]
\begin{center}
\begin{tabular}{r|r|r|r}
& Sentence Pairs & Source Tokens & Target Tokens \\
\hline
French-English        &  2.01m &  62.6m &  56.3m \\
Spanish-English       &  1.97m &  57.1m &  55.0m \\
German-English        &  1.92m &  51.0m &  53.6m \\
\hline
Merged & 11.79m & 335.6m & 335.6m \\
\end{tabular}
\end{center}
\caption{Tokenized corpus statistics for training single-data baselines and $n$-way multilingual models.}
\label{table:tok-corpus-stats}
\end{table}

\begin{table}[t]
\begin{center}
\begin{tabular}{r|c|c|c|c}
& Fr-En & Es-En & De-En & Merged \\
\hline
source (BPE) & 35.3k & 37.2k & 43.1k & 80.0k \\
target (BPE) & 35.3k & 35.2k & 35.1k & 80.0k \\
\end{tabular}
\end{center}
\caption{Source and target vocabulary sizes for single-data baselines and $n$-way multilingual models.}
\label{table:voc-sizes-bpe}
\end{table}

The Europarl evaluation data set dev2006 is used as our validation set, while devtest2006 and test2007 are our blind test sets. We also evaluate the out-of-domain News Commentary test sets nc-dev2007 and nc-devtest2007 in order to demonstrate the robustness of our approach to different kinds of data. Case-sensitive single-reference BLEU scores \cite{papineni:02a} are computed using the \texttt{multi-bleu.perl} script included with Moses. \newcite{DBLP:conf/emnlp/ReimersG17} have shown that reporting a single metric score can sometimes be misleading for many neural architectures. For this reason, all BLEU scores reported in the tables are obtained by averaging the decoding results from five separate models initialized with distinct random seeds.

\subsection{Implementation Details}
Our sequence-to-sequence NMT model with task-specific attention is implemented in C++ using DyNet\footnote{\url{http://dynet.io/}} \cite{dynet}. We use DyNet since the computation graph can be efficiently modified on a batch-by-batch basis during training and decoding, allowing for the runtime selection of attention weights and bias parameters according to the desired task.

We utilize the auto-batching feature \cite{DBLP:conf/nips/NeubigGD17} of DyNet for efficient matrix computations, but the parallel training data must still be separated into batches\footnote{Each batch contains a combined maximum of 5000 source and target tokens.} such that each batch consists of sentences for a single task, e.g. \texttt{<ToEn>} for a batch of sentences that all use the same target-specific attention model for translating into English. The entire training data and batches are shuffled at the beginning of each epoch so that the order of tasks seen during training is random and that they occur in proportion to their distribution in the training data.

We use 256 dimensions for our source and target word embeddings, and 256 dimensions for the hidden states. A single recurrent layer is used for the encoder and decoder. The model parameters are optimized using an unbiased Adam\footnote{We use a learning rate of 0.001.} stochastic optimizer \cite{DBLP:journals/corr/KingmaB14} in order to minimize perplexity on a held-out validation set. The validation set for the single-data baseline systems is simply the dev2006 portion of the official Europarl evaluation data for that language pair, e.g. dev2006.fr and dev2006.en for French-to-English translation. For multilingual systems, we want the model to work well in all possible directions so we merge the dev2006 data for all directions of interest. Combining all six directions gives a validation set with a total of $6 \times 2000 = 12000$ sentences.

\subsection{Task-Specific Attention Model Decoding Results}
Tables \ref{table:bleu-xx-to-en} and \ref{table:bleu-en-to-xx} show decoding results for the Foreign-to-English (\xxen) and English-to-Foreign (\enxx) directions, respectively. In all six translation directions, the single-data baseline systems obtain the highest overall BLEU scores. We expect these baseline models to perform well on their respective test sets since the translation direction (i.e.\ task) of the test set exactly matches the parallel data and validation set used to train the models.

The fully $n$-way multilingual system with a shared attention model shows degradations of up to -2.0 BLEU score, exhibiting a similar pattern to previously reported multilingual NMT results \cite{TACL1081}. When the encoder and decoder both support multiple languages, a single shared attention model is harmful. Our multilingual NMT model with target-specific attention mitigates much of this degradation. We observe gains of between +0.5 and +0.9 BLEU, with respect to the single attention model shared across all language pairs.

Source-specific attention is also better than the shared attention model, but not as good as target-specific attention. Paired attention (i.e.\ a separate set of attention weights and biases for each of the six translation directions with explicitly paired parallel data) shows little change compared to the standard shared attention model. The paired attention model has more tasks than the other models (6 for paired attention vs.\ 4 for both target-specific and source-specific attention). It has therefore seen less data for each task and benefits from no sharing of attention-related parameters. The paired model probably lacks sufficient training data to learn a good separate attention for all tasks.

Table \ref{table:bleu-nc-sets} shows BLEU scores on the out-of-domain News Commentary nc-dev2007 and nc-devtest2007 test sets. Our task-specific attention model again provides gains of between +0.6 to +1.2 BLEU showing that the technique is robust even for test sets less closely matched to the parallel training data.


\begin{table}[t]
\begin{center}
\begin{tabular}{r|ccc|ccc|ccc}
\hline
& \multicolumn{3}{c|}{Fr-En} & \multicolumn{3}{c|}{Es-En} & \multicolumn{3}{c}{De-En} \\
\hline
single-data     & 31.84 & 32.21 & 31.80 & 32.17 & 32.09 & 32.11 & 28.39 & 28.67 & 28.32 \\
\hline
shared          & 29.99 & 30.34 & 30.12 & 30.99 & 30.87 & 30.78 & 26.30 & 26.52 & 26.06 \\
target-specific & \bf30.50 & \bf31.00 & \bf30.68 & \bf31.58 & \bf31.62 & \bf31.63 & \bf26.96 & 27.12 & \bf26.85 \\
source-specific & 30.47 & 30.87 & 30.36 & 31.47 & 31.54 & 31.51 & 26.75 & \bf27.13 & 26.70 \\
paired          & 29.99 & 30.62 & 30.23 & 30.93 & 31.23 & 31.17 & 26.43 & 26.81 & 26.38 \\
\end{tabular}
\caption{BLEU scores for single-data baseline and task-specific attention model variants: \xxen}
\label{table:bleu-xx-to-en}
\vspace{5mm}
\begin{tabular}{r|rrr|rrr|rrr}
\hline
& \multicolumn{3}{c|}{En-Fr} & \multicolumn{3}{c|}{En-Es} & \multicolumn{3}{c}{En-De} \\
\hline
single-data     & 32.24 & 32.44 & 32.87 & 32.60 & 32.60 & 33.18 & 22.15 & 22.55 & 22.36 \\
\hline
shared          & 29.44 & 29.85 & 30.23 & 30.03 & 30.50 & 31.02 & 19.40 & 19.84 & 19.63 \\
target-specific & \bf30.06 & \bf30.38 & \bf30.92 & \bf30.62 & \bf30.93 & \bf31.56 & \bf20.16 & \bf20.51 & \bf20.27 \\
source-specific & 29.68 & 30.20 & 30.50 & 30.54 & 30.70 & 31.48 & 19.66 & 20.10 & 20.01 \\
paired          & 29.19 & 29.68 & 30.07 & 30.17 & 30.40 & 31.07 & 19.62 & 20.13 & 19.97 \\
\end{tabular}
\end{center}
\caption{BLEU scores for single-data baseline and task-specific attention model variants: \enxx}
\label{table:bleu-en-to-xx}
\end{table}

\begin{table}[ht]
\begin{center}
\begin{tabular}{r|rr|rr|rr}
\hline
& \multicolumn{2}{c|}{Fr-En} & \multicolumn{2}{c|}{Es-En} & \multicolumn{2}{c}{De-En} \\
\hline
shared          & 23.20 & 21.56 & 30.00 & 28.31 & 20.60 & 18.40 \\
target-specific & \bf23.82 & \bf22.45 & \bf30.95 & \bf29.47 & \bf21.65 & \bf19.40 \\
\hline
& \multicolumn{2}{c|}{En-Fr} & \multicolumn{2}{c|}{En-Es} & \multicolumn{2}{c}{En-De} \\
\hline
shared          & 26.05 & 24.95 & 31.74 & 30.36 & 16.20 & 14.44 \\
target-specific & \bf26.96 & \bf25.51 & \bf32.64 & \bf30.91 & \bf16.99 & \bf15.12 \\
\end{tabular}
\end{center}
\caption{BLEU scores for single-data baseline and task-specific attention model variants on out-of-domain News Commentary testsets nc-dev2007 and nc-devtest2007.}
\label{table:bleu-nc-sets}
\end{table}

\subsection{Zero-Shot Decoding Results}
Table \ref{table:bleu-zero-shot} shows decoding results for the six zero-shot translation pairs, i.e. those directions (such as French-to-Spanish) never explicitly paired in the parallel data. Although the absolute numbers are lower (as expected for zero-shot pairs), our target-specific attention model gives gains of between +1.0 and +1.5 BLEU over the model that shares attention parameters across all languages. Source-specific attention does not work well for zero-shot translation.

No single-data baseline exists for zero-shot translation since we have no parallel data for those pairs. However, we can pivot by data (i.e.\ find new parallel sentence pairs $x:z$ from existing data $x:y$ and $y:z$, with common $y$) or translate by bridging (translate from $x$ to $y$, and then from $y$ to $z$) for the zero-shot pairs. Pivoting by data is possible since there is a high degree of overlap amongst the various languages of the Europarl Corpus.

Although our multilingual NMT model has not seen explicitly paired data in the zero-shot directions, the encoder and decoder have seen many of the source or target sentences paired with other languages. Our zero-shot experiments are designed to test the hypothesis that the target-specific attention model is better than a model which shares attention parameters for all languages. Any benefit due to source or target data similarity applies equally to each of these multilingual models.

\begin{table}[t]
\begin{center}
\begin{tabular}{r|rrr|rrr|l}
\hline
& \multicolumn{3}{c|}{Fr-Es} & \multicolumn{3}{c|}{Fr-De} & \\
\hline
shared          & 13.64 & 13.63 & 13.75 &  7.82 &  8.04 &  7.84 & \multirow{3}{*}{From Fr} \\
target-specific & \bf15.10 & \bf15.29 & \bf15.31 &  \bf8.76 &  \bf8.95 &  \bf8.82 & \\
source-specific & 12.27 & 12.35 & 12.31 &  6.99 &  7.01 &  6.82 & \\
\hline
& \multicolumn{3}{c|}{Es-Fr} & \multicolumn{3}{c|}{Es-De} & \\
\hline
shared          & 13.43 & 13.33 & 13.48 &  7.57 &  7.82 &  7.57 & \multirow{3}{*}{From Es} \\
target-specific & \bf14.55 & \bf14.40 & \bf14.40 &  \bf8.70 &  \bf8.63 &  \bf8.65 & \\
source-specific & 12.53 & 12.35 & 12.26 &  7.07 &  6.88 &  6.86 & \\
\hline
& \multicolumn{3}{c|}{De-Fr} & \multicolumn{3}{c|}{De-Es} & \\
\hline
shared          & 10.50 & 10.12 & 10.31 &  9.86 & 10.06 & 10.02 & \multirow{3}{*}{From De} \\
target-specific & \bf11.53 & \bf11.24 & \bf11.38 & \bf11.28 & \bf11.31 & \bf11.29 & \\
source-specific &  9.64 &  9.59 &  9.41 &  8.76 &  8.83 &  8.78 & \\
\end{tabular}
\end{center}
\caption{BLEU scores for multilingual NMT `zero-shot' translations comparing the baseline shared attention model to target-specific and source-specific attention models.}
\label{table:bleu-zero-shot}
\end{table}

\section{Analysis and Examples}
\label{sec:analysis-examples}
\begin{figure}[t]
\centering
\includegraphics[width=0.6\textwidth]{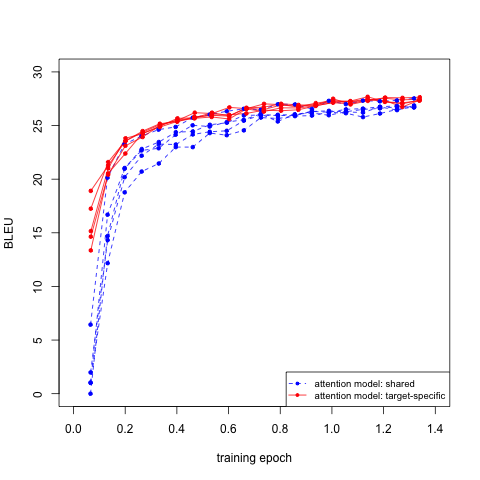}
\caption{Multilingual validation set BLEU scores obtained during training for five different randomly initialized seeds using a shared or target-specific attention model.}
\label{fig:learning-curves}
\end{figure}

Learning curves for the shared attention and target-specific attention model variants are shown in Figure \ref{fig:learning-curves}. The plot includes five curves for each model variant, corresponding to different random initializations. Our target-specific attention model achieves much higher BLEU scores on the validation set in earlier epochs. There is also considerably less variance. The first few times the validation set is decoded with the shared attention model, the gap with respect to the target-specific attention model can be as large as 10 points BLEU. Our validation set contains sentences from all six translation directions so it is not surprising that a model with a single set of shared attention weights and biases performs poorly when applied to the `wrong' task. Even after a full epoch of training (11.8m sentence pairs), the shared attention model continues to lag behind the target-specific attention model by almost one BLEU point.

The following example shows German-to-English translations obtained using a multilingual NMT system with a shared or target-specific attention model. This is an interesting example since the model must attend to the distant German source words ``angesprochen worden ist'' when producing the English translation. The multilingual NMT system with a shared model of attention produces a deficient translation, containing multiple repetitions of the word `report' and omitting the important content word `aid':

\begin{center}
\vspace{0.25cm}
\begin{tabular}{c|p{11cm}}
source & Ich weise auch darauf hin , dass die Frage der Stein@@ koh@@ le sowohl im Wettbewerbs@@ bericht als auch im Beihil@@ fen@@ bericht , über den wir heute diskutieren , angesprochen worden ist . \\
\hline
reference & I would also draw your attention to the fact that the coal issue is raised both in the competition report and in the subsidy report that we are discussing today . \\
\hline
shared & I also note that the issue of coal is raised by the coal report , both in the report on the competitiveness report and on the report on which we are debating today . \\
\hline
target-specific & I would also point out that the issue of coal has been raised both in the competition report and in the aid report we are discussing today . \\
\end{tabular}
\vspace{0.25cm}
\end{center}

\begin{figure}[t]
\begin{subfigure}{.52\textwidth}
\centering
\includegraphics[width=\textwidth]{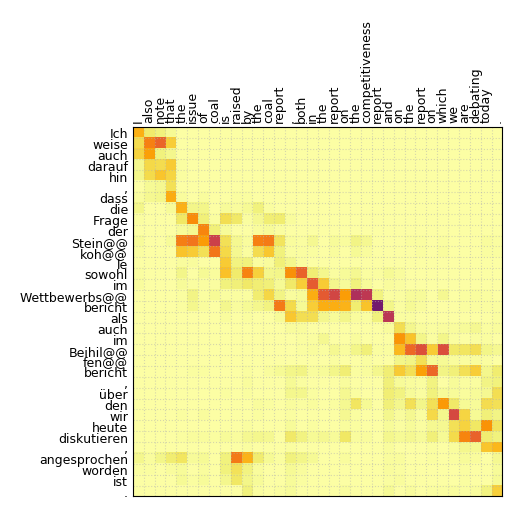} 
\caption{Attention model: shared}
\label{fig:attention-plots-shared}
\end{subfigure}
\begin{subfigure}{.48\textwidth}
\centering
\includegraphics[width=\textwidth]{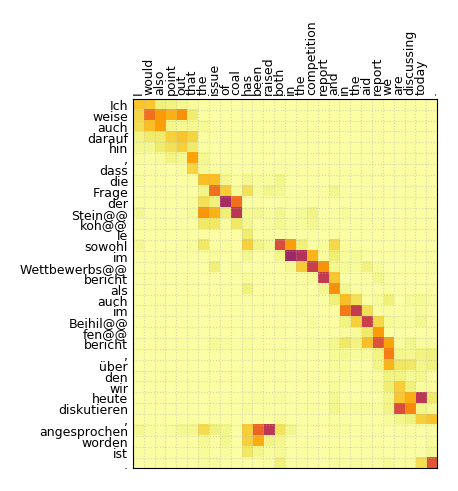} 
\caption{Attention model: target-specific}
\label{fig:attention-plots-target}
\end{subfigure}
\caption{Attention plots obtained while decoding a German-to-English sentence pair in the test set. Rows correspond to source words and columns to each target position. Target-specific attention results in less diffuse alignments.}
\label{fig:attention-plots}
\end{figure}

The attention matrices computed during decoding for the shared and target-specific model variants are shown in Figure \ref{fig:attention-plots}. Rows correspond to the words of the German source sentence and columns to the words of the English target translation. The target-specific attention model results in a sharper and less diffuse attention over the words of the source sentence, especially at the beginning and end of the sentence, and for the translation of the passive German construction ``angesprochen worden ist'' which requires long-distance re-ordering.

For this example, the shared attention model leads to diffuse alignments since a single set of attention weights and biases must be used to align the words of many target languages with disparate word orders. Our target-specific attention model leads to more accurate alignments since the decoder is conditioned for a single target language (as indicated by the `task' token prefix). Given the target-specific set of attention weights and biases, and the initialization of the decoder state using the RNN encoded source, the decoder is able to more accurately attend to the correct source words when producing the words of the target sentence at each time-step.

\section{Conclusions and Future Work}
\label{sec:conclusions-future-work}
We have described a simple but effective technique for improving the quality of multilingual NMT. Our technique mitigates much of the loss that occurs when multiple source and target languages are supported in a system with a single encoder and decoder. It is particularly effective in the zero-shot (i.e.\ extreme low-resource) translation directions.

Our approach extends the use of target language prefix tokens described by \newcite{TACL1081} to select a task-specific set of attention weights and biases for each task of interest. For NMT, where attention can be considered an analogue of word alignment, the use of separate attention weights and biases for language pairs with different word orders leads to improved BLEU scores in our multilingual decoder. Our results show that multilingual NMT works best with a target-specific attention model, i.e.\ a distinct set of attention weights and bias parameters for each supported target language. Our improved model handles all possible translation directions with only a small increase in the total number of parameters, compared to the single-data baseline or standard shared attention model systems.

In future work, we plan to apply our multilingual techniques to true low-resource language pairs, with the goal of augmenting low-resource poor quality translation systems with knowledge obtained on languages with richer resources. Given our results for zero-shot translation, we expect our approach to work well. Our attention model variants can also be applied to multi-task learning frameworks, e.g.\ to improve the quality of translation using knowledge learned from a variety of other natural language processing tasks such as POS tagging and dependency parsing \cite{DBLP:journals/tacl/KiperwasserB18}. Recent work on learning and unsupervised induction of multilingual word representations \cite{DBLP:journals/corr/UpadhyayFDR16,zhang-EtAl:2017:Long5} could also be used to improve multilingual translation since the models share a single vocabulary for all source and target languages of interest.


\bibliographystyle{acl}
\bibliography{biblio}

\end{document}